# A Self-Constructing Multi-Expert Fuzzy System for High-dimensional Data Classification

Yingtao Ren, Yu-Cheng Chang, Thomas Do, Zehong Cao, *Senior Member, IEEE,*
and Chin-Teng Lin, *Fellow, IEEE*

*Abstract*—**Fuzzy Neural Networks (FNNs) are effective machine learning models for classification tasks, commonly based on the Takagi-Sugeno-Kang (TSK) fuzzy system. However, when faced with high-dimensional data, especially with noise, FNNs encounter challenges such as vanishing gradients, excessive fuzzy rules, and limited access to prior knowledge. To address these challenges, we propose a novel fuzzy system, the Self-Constructing Multi-Expert Fuzzy System (SOME-FS). It combines two learning strategies: mixed structure learning and multi-expert advanced learning. The former enables each base classifier to effectively determine its structure without requiring prior knowledge, while the latter tackles the issue of vanishing gradients by enabling each rule to focus on its local region, thereby enhancing the robustness of the fuzzy classifiers. The overall ensemble architecture enhances the stability and prediction performance of the fuzzy system. Our experimental results demonstrate that the proposed SOME-FS is effective in high-dimensional tabular data, especially in dealing with uncertainty. Moreover, our stable rule mining process can identify concise and core rules learned by the SOME-FS.**

Fuzzy neural networks (FNNs) succeeded in many real-world applications, with TSK fuzzy systems [1] being the most widely used. Ongoing advancements have yielded a range of optimization strategies for FNNs. Notably, for network structure learning, approaches such as ANFIS [2], SONFIN [3], and various genetic algorithm-based rule learning methods have been introduced [4]–[7]. In parameter adjustment, prevalent methods include gradient descent-based and least squares estimation-based methods, which are commonly employed [8], [9]. These optimization strategies enhance the efficiency and accuracy of FNNs, making them more suitable for real-world problems.

However, as the dimensionality and quantity of data increase, the prediction performance of FNNs gradually declines [10]–[12]. Furthermore, in real-world classification tasks, the increase in data dimensionality is often accompanied by increased data noise, requiring models to have stronger capabilities for handling noisy data. Meanwhile, TSK fuzzy classifiers tend to become overly complex and cannot even compute the results when handling high-dimensional data [13]. Many previous works have used clustering algorithms or feature dimensionality reduction methods to preprocess high-dimensional data before feeding it into fuzzy neural networks. For example, principal component analysis or clustering-based dimensionality reduction methods [13]–[16] can be used to enhance fuzzy classifiers, effectively mitigating the adverse effects of high-dimensional inputs on fuzzy systems. However, these methods not only result in excessive loss of original information during data preprocessing but also make the generated fuzzy rules more difficult to interpret [17], [18], leading to a significant decrease in both accuracy and interpretability.

With the prospective development in deep learning (DL), many algorithms extend components from DL to enhance the ability of fuzzy systems in high-dimensional data and alleviate vanishing gradients. For example, Wu *et al*. [20] use mini-batch gradient descent and regularization to train weights effectively. Cui *et al*. [21], [22] integrate layer normalization to transform firing strength values, which expedites the convergence. Zhang *et al*. [23] modify the calculation of antecedents and the propagation of gradients to improve the training efficiency of antecedents. However, these methods based on individual TSK models still have difficulty achieving satisfactory performance in high-dimensional data, especially in datasets with uncertainty [23].

In classic machine learning, multiple ensemble techniques (Bagging *et al.*) are commonly used to compensate for the limitations of individual classifiers [25]–[28]. Among these, many ensemble methods based on decision trees (DT) are still the best way to handle typical tabular data [28], especially when dealing with noisy and high-dimensional data. Both DT and fuzzy systems use rules to make predictions [29]. Additionally, fuzzy rules are often more interpretable and flexible than crisp rules [30]. FNNs utilize linguistic rules to capture complex patterns in data [31], making them well-suited for use as base learners. Using FNNs as base classifiers also presents certain challenges, such as the difficulty in tuning parameters and insufficient internal specificity. These

This work was supported in part by Australian Research Council (ARC) under Discovery Grant DP210101093 and Grant DP220100803, in part by the University of Technology Sydney (UTS) Human-Centric Artificial Intelligence (AI) Centre funding sponsored by GrapheneX (2023–2031), in part by Australia Defence Innovation Hub under Contract P18-650825, in part by Australian Cooperative Research Centres Projects (CRCP) Round 11 under Grant CRCPXI000007, in part by the U.S. Office of Naval Research Global under Cooperative Agreement ONRG-NICOP-N62909-19-1-2058, in part by the Air Force Office of Scientific Research (AFOSR)—Defence Science and Technology (DST) Australian Autonomy Initiative under Agreement ID10134, in part by the New South Wales (NSW) Defence Innovation Network and the NSW State Government of Australia under Grant DINPP2019 S1-03/09 and Grant PP21-22.03.02.

Yingtao Ren, Yu-Cheng Chang, Thomas Do and Chin-Teng Lin are with University of Technology Sydney, Ultimo, NSW 2007, Australia (e-mail: yingtao.ren@student.uts.edu.au; Yu-Cheng.Chang@uts.edu.au; thomas.do@uts.edu.au and Chin-Teng.Lin@uts.edu.au). Zehong Cao is with STEM, University of South Australia, Adelaide, SA 5000, Australia (e-mail: jimmy.cao@unisa.edu.au). (Corresponding author: Chin-Teng Lin.)

issues can limit the effectiveness and applicability of FNNs in complex scenarios.

Inspired by the success of individual TSK models and the promising prospects of combining multiple FNNs into an effective fuzzy system, this study focuses on how to learn and combine several fuzzy learners to better fit high-dimensional data. Building upon the strengths of individual FNNs, we introduce an innovative approach by combining multiple FNNs into an effective fuzzy system named Self-Constructing Multi-Expert Fuzzy System. This not only improves the ability to model high-dimensional data but also enhances the robustness of the system.

Within the proposed method, each fuzzy rule can be regarded as a local expert, and a base classifier is viewed as a group of experts [29]. Within each base classifier, the various fuzzy rules operate in parallel, focusing more on their respective local regions, effectively avoiding the vanishing gradient problem. Each base learner develops a unique structure within its local region and automatically selects the appropriate fuzzy rule for decision-making based on the characteristics of the input samples. Finally, we design a specific rule-based voting mechanism to integrate the results, providing more diversity and stability in the decision-making process.

In the aspect of ensemble methods [27], SOME-FS offers inherent ease of hyperparameter adjustment, facilitating straightforward tuning and optimization. Simultaneously, it achieves strong diversity characteristics by incorporating a variety of structures of models. This balance enables SOME-FS to optimize performance effectively, uniting straightforward hyperparameter tuning with robust model diversity to enhance overall system effectiveness.

Detailed experiments demonstrate that the proposed method exhibits strong noise resistance and maintains unique interpretability on high-dimensional data compared to other state-of-the-art classifiers.

Our main contributions are as follows:

- We design a novel and effective self-constructing multi-expert fuzzy system that is highly effective in handling high-dimensional tabular data, particularly when dealing with uncertainty. Additionally, it requires very few hyperparameters to tune, making it highly suitable for complex classification tasks.
- We propose a mixed self-constructing learning method that introduces significant variation in the number of fuzzy rules and membership functions across different base classifiers, which enhances robustness under various data conditions and requires very few hyperparameters to tune.
- We propose a multi-expert advanced strategy, combining rule selection and rule-based voting to integrate fuzzy rules within the ensemble architecture effectively. This strategy extracts the most prevalent rules to make decisions directly, greatly enhancing interpretability and stability when dealing with high-dimensional data.

The remainder of this article is organized as follows. Section II describes related work. Section III introduces the details of our proposed SOME-FS, and Section IV validates the performance of SOME-FS on high-dimensional datasets with varying noise levels. Section V draws conclusions and points out some future research directions.

## II. RELATED WORK

### A. Self-constructing Neural Fuzzy Inference Network

The TSK fuzzy system has been widely researched for its universal approximation ability, robustness, and suitability in various practical applications, especially in process control and expert systems. The Online Self-Constructing Neural Fuzzy Inference Network (SONFIN) is a classic first-order TSK fuzzy model [3], [33], [34]. It employs self-constructing learning and parameter learning, thereby avoiding the need for pre-defined membership functions. The structure learning process involves both the precondition and consequent structure identification of a fuzzy IF–THEN rule which is described as:

$$Rule_k : \text{IF } x_1 \text{ is } A_{k,1} \text{ and } \dots \text{ and } x_D \text{ is } A_{k,D},$$

$$\text{THEN } y_k^1(x) = b_{k,0}^1 + \sum_{d=1}^{D} b_{k,d}^1 \cdot x_d \text{ and } \dots$$

$$\text{and } y_k^C(x) = b_{k,0}^C + \sum_{d=1}^{D} b_{k,d}^C \cdot x_d \quad (1)$$

in which $A_{k,1}$ ($k = 1, \dots, K; d = 1, \dots, D$) is the membership function (MF) for the $d$-th antecedent in the $k$-th rule, $b_{k,0}^1$ and $b_{k,d}^C$ ($c = 1, \dots, C$) are the consequent parameters for the $c$-th class in the classification task. We will describe the specific definitions of six layers in the SONFIN.

#### 1) Layer 1 (Input Layer):

There are no computations performed in this layer. Each node in this layer corresponds to an input variable and solely transmits the input values directly to the next layer.

$$O_d^1 = x^d \quad (2)$$

#### 2) Layer 2 (Membership Layer):

Each node in this layer corresponds to one linguistic label (such as small, large, etc.) of an input variable in Layer 1. Layer 2 calculates the membership value, which specifies the degree to which an input value belongs to a fuzzy set. Various types of membership functions (MFs) can be utilized in SONFIN, including triangular, trapezoidal, and Gaussian, provided these are differentiable. For simplicity, this article focuses on Gaussian MFs, and the membership grade of $x_d$ on $A_{k,d}$ is

$$O_{k,d}^2 = u_{k,d}(x_d) = \exp\left(-\frac{(x_d - m_{k,d})^2}{2\sigma_{k,d}^2}\right) \quad (3)$$

where $m_{k,d}$ and $\sigma_{k,d}^2$ are the center and standard deviation of the Gaussian MF, respectively.

#### 3) Layer 3 (Rule Layer)

A node in this layer represents a fuzzy logic rule and performs precondition matching. For each Layer 3 node, we employ the following AND operation

$$O_k^3 = f_k = \prod_{d=1}^{D} u_{k,d}(x_d), \qquad (4)$$

$f_k$ is the firing strength of Rule k. In the training process, for each input data cluster $\vec{x}$, find

$$J = arg \max_{1 \le j \le r(t)} f_k \qquad (5)$$

If $J < \Phi_{in}$, generate a new rule, set r(t) = r(t) +1, set $m$ and $\sigma$ as the center point and variance of the cluster. $\Phi_{in}$ is a pre-defined threshold ($0 < \Phi_{in} < 1$), r(t) is the number of fuzzy rules. In other words, there are no rules initially in SONFIN. Rules are created and adapted as online learning progresses through simultaneous structure and parameter identification.

### 4) Layer4 (Normalization Layer):

The number of nodes in this layer matches that of Layer 3, and the firing strength calculated in Layer 3 is normalized here by

$$O_k^4 = \bar{f}_k = f_k / \sum_k f_k, \qquad (6)$$

$\bar{f}_k$ is the normalized firing strength of the k-th rules.

### 5) Layer 5 (Consequent Layer)

This layer is called the consequent layer. For classification tasks, the rule output should be a vector representing the probabilities of belonging to different classes. The calculation process within each node is as follows equation

$$O_k^5 = y_k = \bar{f}_k (c_{k,0} + c_{k,1}x_1 + \ldots + c_{k,n}x_n), \qquad (7)$$

In which $y_k$ represents the output of each fuzzy rule, which is the product of the normalized firing strength and the polynomial of the input features. $c_{k,n}$ is a C-dimensional vector in a C-classes classification task, and the SoftMax function is applied to map this vector to one representing probability.

### 6) Layer 6 (Summation Layer):

Each node in this layer corresponds to a single output variable. The node integrates all the actions recommended by Layer 5 and functions as a defuzzifier with

$$O^5 = \sum_k y_k. \qquad (8)$$

SONFIN dynamically grows its network structure during learning, optimizing for both economy and speed, which generates fewer fuzzy rules for simpler models or enhances accuracy. However, it struggles with large datasets and numerous features, leading to poor performance and reduced interpretability due to its complex adaptive mechanisms.

### B. Ensemble Learning and Fuzzy Ensemble System

Ensemble learning methods combine multiple machine learning models to improve predictive performance. Their central idea is that a group of weak learners can collectively form a strong learner. The main advantages of ensemble learning are its ability to reduce overfitting, enhance robustness, and improve accuracy. Ensemble learning is particularly effective in scenarios where a single model's performance is limited. Common methods include bagging, boosting, and stacking [24]–[26], each utilizing different strategies to train and combine models. Bagging is one of the most used ensemble strategies, which involves training different classifiers using bootstrapped replicas of the original training dataset. Specifically, a new dataset is formed for each classifier by randomly drawing instances from the original dataset with replacement, typically maintaining the original dataset size. This resampling procedure ensures diversity by utilizing different data subsets, which is crucial for the effectiveness of ensemble methods [35]. In the advanced bagging method known as random forest (RF), each classifier is assigned a different set of features, further enhancing the diversity among classifiers. Finally, when an unknown instance is presented to each individual classifier, the class is inferred using a majority or weighted vote.

Compared to ensemble methods that use traditional (crisp) classifiers such as decision trees and neural networks [30], applying fuzzy classifiers as the base classifier can significantly enhance the robustness and expand the scope of applications [36]. These approaches enhance parallel or incremental learning and maintain high interpretability. Various fuzzy ensemble systems have been developed, including bagging or boosting fuzzy rule-based models [37], fuzzy support vector machine ensembles (E-SVM) [38], and fuzzily weighted AdaBoost ensembles (FWAdaBoost) [39]. Fuzzy ensemble systems have been widely applied in practical applications, including wind speed forecasting, fault diagnosis and disease diagnosis [40]–[42].

## III. Method

This section presents the proposed SOME-FS method and its interpretation strategy, whose overall architecture is shown in Figure 1. The section includes three sub-sections: (1) the base learner EA-SONFIN, which serves as the foundational element of our approach; (2) the adaptive multi-expert learning architecture of SOME-FS, which enhances the predictive performance through collaborative learning among multiple experts; and (3) specialized a stable rule extraction method that contributes to the interpretability of the model by extracting short and meaningful stable rules.

### A. Single EA-SONFIN Construction and Training

We observe that self-constructing TSK classifiers, such as SONFIN, generated through structure learning meet the requirements of ensemble learning theory for base learners effectively. Moreover, the potential feature subset segmentation of the bagging method also alleviates the limitations of TSK on high-dimensional data.

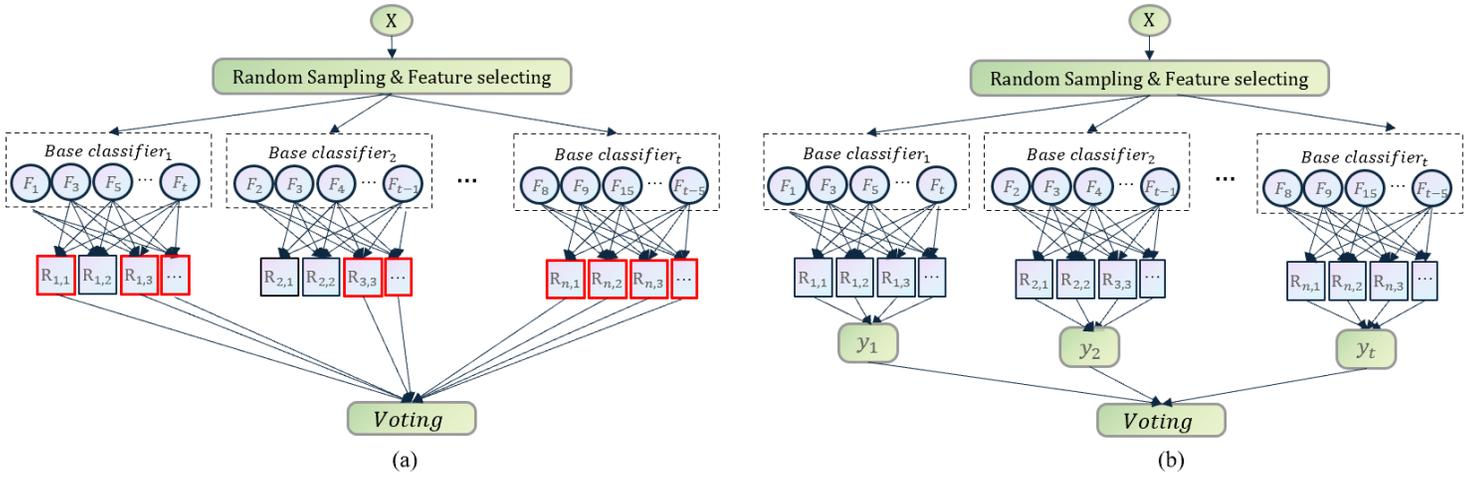

Fig. 1. Overall architecture of (a) SOME-FS (b) Classic Bagging-TSK

Inspired by recent advancements in deep learning architecture and multi-expert learning, we innovatively propose the EA-SONFIN base classifier, expanding the advantages of fuzzy rules to a kind of multi-expert decision. The subsequent ablation experiment also confirmed its effectiveness. Subsequently in this subsection, we will provide a detailed introduction to the EA-SONFIN base classifier.

The structure of each EA-SONFIN is determined by the coverage of the rules over the input feature space, following the methods in articles [2], [3]. However, unlike these methods, we combine partition-based fuzzy rule initialization and density-based online rule learning to a mixed self-constructing learning. We generate an overall fuzzy antecedent at first, then use self-constructing learning to supplement the fuzzy rules, balancing the global and local data characteristics.

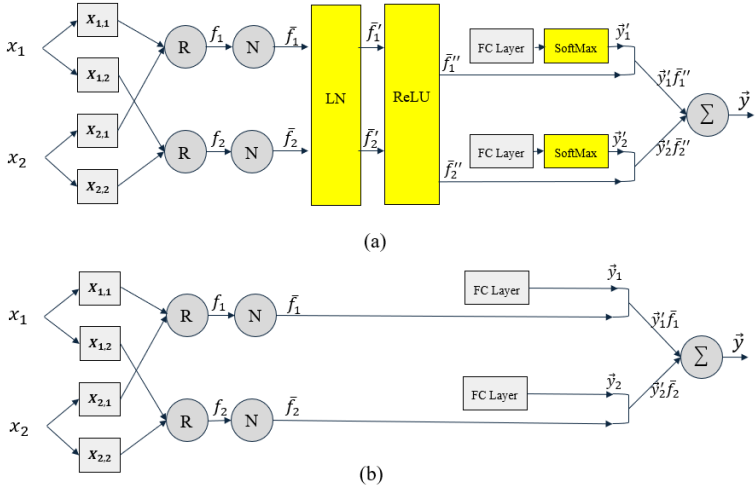

Fig. 2. Structure of our proposed (a)EA-SONFIN and (b) classic TSK.

Firstly, we use a partition-based clustering method to identify $p$ partitions for constructing the initial antecedent fuzzy sets. Supposing that the centers of $p$ partitions are

$$\vec{P}(i) = [\vec{x}_1(i), \ldots, \vec{x}_m(i)], \quad i = 1, \ldots, p. \quad (9)$$

Each clustering partition is used to generate a fuzzy rule. Each rule is generated with the centers and widths of the fuzzy antecedent set as follows,

$$\vec{c}_i^1 = \vec{P}(i), \sigma_i^1 = \sigma_{init}, i = 1, \ldots p \quad (10)$$

where $\vec{c}_i^p$ and $\vec{\sigma}_p$ are the centers and widths of the fuzzy antecedent of the $p$-th fuzzy rule.

Secondly, we perform density-based clustering on each training batch to obtain cluster centers and variances. We use these cluster centers for self-constructing learning, calculating the maximum activation $J$ using formula (5) and similar value S [44], and comparing them with the pre-defined threshold. The specific process is detailed in Table I.

Adding too many modules to a single FNN will reduce the model's robustness and is prone to overfitting, which is demonstrated in Section IV.B. Therefore, we only add an LN-ReLU component after the firing strength layer to reduce the number of rules to be optimized and to strengthen the focus of each expert on its respective local region. Each fuzzy rule represents the pattern of a local region. The structure of single EA-SONFIN can be seen in Figure 2. Layer Normalization and ReLU are crucial techniques in deep learning models. Layer Normalization stabilizes the firing strengths process by normalizing and scaling the inputs of each layer, as described by the following formula:

$$\overline{f_k}' = \frac{\overline{f_k} - \mu}{\sqrt{\sigma^2 + \epsilon}} \quad (11)$$

where $\overline{f_k}$ is the $k$-th rule's normalized firing strength after layer 4, $\mu$ and $\sigma^2$ are the mean and standard deviation of the $\overline{f_k}$, and $\epsilon$ is a small constant to prevent division by zero.

ReLU is defined by the formula:

$$\overline{f_k}'' = max(0, \overline{f_k}') \quad (12)$$

**Algorithm 1:** Mixed self-constructing learning algorithm for single EA-SONFIN classifier.

For input, we have a training dataset $D$, a maximum number of fuzzy rules $rule_{max}$ and a pre-defined threshold $\Phi_{in}$.

**Begin**

1: Initialize fuzzy rules using the elbow method to determine initial clustering points and related variances.

2: Begin main loop:

3:     Perform training on the existing structure using dataset D.

4:     Apply density-based clustering to identify non-noise center points in each batch.

5:     Calculate maximum activation J and similar value S for each cluster point.

6:     For each cluster point:

7:         If J > $\Phi_{in}$ and S < 0.5 then:

8:             Generate a new fuzzy rule.

9:         End If

10:     End For

11:     If the number of fuzzy rules exceeds $rule_{max}$ or no new rule is learned for three consecutive epochs then:

12:         Exit main loop.

13:     End If

14:     Go to Step 3.

15: End main loop

16: Transition to the pure parameter learning phase.

17: Return trained classifier

**End**

---

Adding a SoftMax function for each fuzzy rule in the consequent layer is represented by the following equation. This allows each fuzzy rule to make decisions individually.

$$\vec{y}_k = SoftMax(c_{k,0} + c_{k,1}x_1 + \ldots + c_{k,n}x_n) \quad (13)$$

Thus, the final output of a single EA-SONFIN classifier is:

$$\vec{y} = \sum_{k=1}^{r} \overline{f_k}'' \vec{y}_k \quad (14)$$

These added components keep the firing strength as interpretable as possible and increase the connections between the linear neural networks within the same classifier. The newly added functions enhance the focus of each fuzzy rule on its local region, enabling the ensemble system to better fit high-dimensional data.

### B. Novel Multi-Expert Fuzzy System Architecture

Despite the TSK classifier's superior interpretability compared to crisp neural networks [39] [44], it faces two inherent challenges when applied to high-dimensional datasets. First, as dimensionality increases, the firing strength values diminish, preventing the antecedents from being adequately trained.

Second, a large number of rules (typically more than 30) are required to fit a high-dimensional feature space, leading to significant overlap, redundant learning in certain local regions, and exacerbating overfitting issues. Bagging can mitigate these issues by allowing other weak learners to correct the final output, even if one weak learner makes a wrong prediction. Meanwhile, each weak learner only focuses on the part of the region, alleviating the high-dimensional curve for FNNs. Hence, designing an effective bagging-based fuzzy system can substantially reduce the complexity of a single classifier and the number of input features to enhance the performance of the traditional fuzzy system. Meanwhile, the abundant diversity introduced by different fuzzy rules is well-suited to bagging's core principle: maximizing the differences among base learners.

Based on the high compatibility between ensemble learning and fuzzy systems, we propose three novel methods that leverage the strengths of both approaches to maximize diversity among fuzzy base learners. Random sampling is performed in accordance with the standard bagging technique. The overall structure of SOME-FS can be seen in Figure 1. The red box represents the selected rules in a decision. Firstly, compared to the traditional voting mode based on the decision results of each base classifier, we adopt a rule-based voting mode that utilizes every single rule within each fuzzy classifier. This approach can more fully exploit the diversity among base classifiers. Secondly, by applying a layer normalization mechanism in each base classifier, the system adjusts the contribution of each fuzzy rule during prediction, effectively avoiding overfitting. Finally, since each base classifier retains a portion of out-of-bag data that is not involved in training, using this data to evaluate the performance of each base classifier and assigning it a performance weight can strengthen the generalization ability of the overall model. The weight of the t-th base classifier is calculated as follows:

$$w_t = \begin{cases} 1, & \text{if } err_t \leq (emin+marg) \\ \frac{(emax+marg)-err_t}{(emax-emin)}, & \text{if } (emin+marg) \leq err_t \leq (emax) \end{cases} \quad (15)$$

where the $err_t$ is the error rate of the t-th base classifier in SOME-FS system. The emax is the maximum error rate in single EA-SONFIN classifiers, which is calculated as:

$$emax = \max\{err_t\} \quad (16)$$

Similarly, the emin is the minimum error rate in the system and $marg = (emax - emin)/4$. This strategy is also used in other fuzzy ensemble learning systems [35], [45].

Thus, the systematic mixed voting that makes the final decision can be represented by the following formula:

$$\hat{y} = \sum_{i=0}^{t} w_i \left(\sum_{k=1}^{r} \overline{f_{i,k}}'' \vec{y}_{i,k}\right) \quad (17)$$

Individual classifiers within the ensemble influence each other, learning different patterns and contributing to the diversity of the system. The number of rules and membership functions also vary significantly across different classifiers, greatly enhancing the system's diversity and randomness. As a result, the entire system can also be viewed as a large mixture of experts (MoE) model, where each expert specializes in different aspects of the input space, providing a comprehensive and robust decision to the classification task.

### C. Interpretable Stable Rule Extraction from Multi-Expert Fuzzy Ensemble Systems

Fuzzy classifiers employ linguistic methods to provide explanations for classification results [46]. However, in high-dimensional classification tasks (involving hundreds of dimensions), the explanations often include numerous semantic features, making it difficult for decision-makers to effectively identify the key factors relevant to the decision from these features. This complexity is a major reason for the reduced transparency of traditional TSK classifiers in high-dimensional problems.

The proposed SOME-FS exhibits significant diversity among different base learners, with each base learner making independent decisions. In this scenario, each of them learns a set of fuzzy rules, and the intersection of these rules forms a set of small and stable rules, providing efficient and transparent explanations. Similar explanation methods have already been used in tree-based ensemble learning systems [47], [48] and can be adapted to SONFIN-based ensemble systems.

The proposed fuzzy TSK-based Interpretable Stable Rule Extraction method consists of three main steps:

1) Rule Generation: Multiple base EA-SONFIN models are employed to generate a diverse set of fuzzy rules. Each base model independently learns rules that capture different aspects of the data.

2) Rule Extraction: Association rule mining algorithms, such as FP-Growth, are applied to extract frequently occurring short and stable rules from these base models. This step identifies rules that are common across different models, indicating robust and significant patterns in the data.

3) Rule Validation: The firing strength is utilized to evaluate the significance of these extracted rules. By calculating the firing strength of each rule across the dataset, we assess their effectiveness and stability, retaining those with higher significance.

In the stage of rule generation, we train multiple base fuzzy models on the dataset. Each of these models operates independently and learns a set of fuzzy rules. These rules typically describe certain patterns and relationships within the data, capturing the underlying structure in a high-dimensional feature space. The independence of each model ensures diversity in the rules learned. During the training phase, each fuzzy model generates rules in the form of "if-then" statements. For instance, a rule might be: "If the temperature is high and humidity is low, then the output is medium." These rules are based on the fuzzy partitions of the input feature space, allowing for smooth transitions between different regions of the input space. The TSK model's fuzzy rules provide a powerful way to model complex nonlinear relationships within the data.

In the rule extraction stage, we apply the FP-Growth association rule mining algorithm, which is well-suited for identifying frequently occurring patterns within a large set of rules. Our goal is to find rules that appear frequently across different base models, indicating that they capture robust and significant patterns in the data. The association rule mining process involves scanning the generated rules and counting their occurrences. We focus on short rules with fewer conditions because they are typically easier to interpret and more likely to represent fundamental relationships in the data. By setting a threshold for the minimum support (frequency of occurrence), we filter out less significant rules and retain only those that are commonly observed. Additionally, during the mining process, we consider only classifiers that make correct predictions and the rules that are activated and yield correct prediction results. This approach enhances the local interpretability [49] of rule extraction by focusing on specific instances where the model performs accurately, allowing us to understand the key factors influencing individual decisions.

The frequent occurrence of such rules suggests that they are capturing essential patterns that are consistently observed across different classifiers. Table X provides a stable rule extraction example. The extracted stable rule is:

*Stable Rule*: IF $x_3$ is high and $x_5$ is high and $x_9$ is low
THEN $y$ is positive

To filter the extracted rules, we can use the firing strength as a basic metric. Firing strength is a measure of how well a fuzzy rule matches the input data. For each extracted stable rule, we calculate its firing strength across the dataset to evaluate its effectiveness and stability. We use the firing strength to rank the extracted rules. Rules with higher firing strengths are considered more significant as they have a stronger presence in the dataset. This step ensures that the rules we extract are not only frequent but also robust and meaningful.

TABLE X
A SAMPLE OF STABLE RULE EXTRACTION (STABLE RULES REPRESENT CONCISE AND IMPORTANT PATTERNS)

|  | Rule 1 | Rule 2 | Rule 3 | Stable Rule |
|---|---|---|---|---|
| x1 | high | very high | low |  |
| x2 | very low | very high | very low |  |
| x3 | **high** | **high** | **high** | **high** |
| x4 | med | low | low |  |
| x5 | high | med | very low |  |
| x6 | **very high** | **very high** | **very high** | **very high** |
| x7 | med | low | med |  |
| x8 | high | very low | low |  |
| x9 | med | low | low |  |
| x10 | **low** | **low** | **low** | **low** |
| x11 | very low | very low | low |  |
| y | positive | positive | positive | positive |

## IV. Experiment and Result

In this section, we evaluate the performance of our proposed SOME-FS across multiple datasets from various application domains, focusing on high-dimensional tabular data.

We evaluated all our algorithms on 12 classification datasets from public machine learning repositories, with their characteristics summarized in Table II. We performed five-fold cross-validation on each dataset and reported the average performance.

TABLE II
SUMMARY OF THE TWELVE DATASETS

| Num | Dataset | Dimensions | Samples | Class |
|---|---|---|---|---|
| 1 | Parkinson's Disease[1] | 754 | 756 | 2 |
| 2 | Gas Sensor Array Drift[1](GSAD) | 128 | 13910 | 6 |
| 3 | Musk[1] | 166 | 6598 | 2 |
| 4 | Isolet[1] | 617 | 7797 | 26 |
| 5 | Secom[1] | 591 | 1567 | 2 |
| 6 | Human Activity[1] | 561 | 10299 | 6 |
| 7 | Spambase[1] | 57 | 4601 | 2 |
| 8 | Mice Protein Expression[1] | 80 | 1080 | 8 |
| 9 | Kitchen[2] | 300 | 8322 | 2 |
| 10 | Book[2] | 300 | 35524 | 2 |
| 11 | Movie[2] | 300 | 13420 | 2 |
| 12 | Music[2] | 300 | 1358 | 2 |

[1]https://archive.ics.uci.edu/datasets?Task=Classification&skip=0&take=10&sort=desc&orderBy=NumHits&search=
[2]https://cseweb.ucsd.edu/~jmcauley/datasets.html#amazon_reviews

In this section, we compare seven classifiers to evaluate our proposed approach. Among them are three state-of-the-art fuzzy classifiers (HTSK-LN-ReLU [50], TSK-MBGD [19] and RFNN [21]), one basic TSK classifiers SONFIN [3], one widely used tree-based bagging algorithm (Random Forest [51]), decision tree and our proposed SOME-FS. This selection covers various types of fuzzy classifiers and widely used bagging methods, providing a comprehensive performance evaluation.

### A. General Performance

In this section, we evaluate the classification performance of the proposed algorithm on twelve datasets. We performed thirty rounds of grid search cross-validation on each compared algorithm to fine-tune the optimal hyperparameters. For SOME-FS, the initial number of structural clusters was set to 5, and the maximum number of rules was capped at 30. The number of base learners was set to 10. In the structure learning process, we choose the k-means as the partition-based clustering method and DBSCAN as the density-based clustering method.

In the experiments, we conducted five-fold cross-validation for each algorithm using the same random seed. Classification accuracy was used to evaluate the performance of each algorithm. The detailed experimental results are summarized in Table III. Table III shows that SOME-FS and RF stand out due to their consistent and robust performance across several datasets. The performance of SOME-FS ranks first among all evaluated algorithms. Additionally, SOME-FS excels in six datasets, while RF excels in four. Meanwhile, RFNN demonstrates strong capabilities in complex pattern recognition, which is evident from its performance in two datasets.

To further explore the differences between the results of each classifier, we performed a nonparametric test on the accuracy score. We chose the Friedman test [52], which is widely used for comparing algorithm performance. Furthermore, we set the proposed SOME-FS as the control classifier and conduct pairwise comparisons with all other classifiers, adjusting the p-values using the Holm post hoc test. The results are reported in Table V and Table VI. For all hypothesis tests, we set the significance level (p-value) at 0.05.

The Friedman test produces a p-value $< 0.01$, which indicates significant differences among the tested algorithms. Subsequent post hoc tests revealed no statistically significant difference between the proposed algorithm and RFNN or RF, but significant differences were observed when compared to the other four algorithms. Among the seven tested algorithms, SOME-FS achieved the highest mean rank scores, followed by RFNN and RF. It can be concluded that SOME-FS demonstrates competitive performance and is an effective classifier for high-dimensional tabular datasets.

TABLE III
ACCURACY OF DIFFERENT METHODS ON VALIDATION NO NOISE DATASETS

| No Noise | Parkinson Disease | GSAD | Musk | ISOLET | Secom | Human Activity | Spambase | Protein | Kitchen | Book | Movie | Music |
|---|---|---|---|---|---|---|---|---|---|---|---|---|
| **SOME-FS** | **88.49** | **99.14** | **99.42** | **93.84** | 92.15 | 89.20 | 93.63 | 98.70 | 56.36 | **66.28** | **72.27** | 62.00 |
| HTSK-LN-ReLU | 86.37 | 97.46 | 97.89 | 85.24 | 88.90 | 87.19 | 93.31 | 96.67 | 54.64 | 61.58 | 65.22 | 58.02 |
| RF | 86.77 | 98.53 | 97.57 | 89.74 | **93.30** | **91.30** | **94.87** | 97.59 | 56.90 | 61.84 | 71.48 | **68.12** |
| TSK-MBGD | 83.99 | 98.38 | 96.94 | 88.20 | 88.19 | 85.35 | 92.98 | 95.19 | 55.29 | 61.82 | 65.76 | 60.53 |
| SONFIN-BN | 81.74 | 98.48 | 97.29 | 87.11 | 88.70 | 82.96 | 93.57 | 95.46 | 54.68 | 63.26 | 70.54 | 61.44 |
| RFNN | 87.16 | 99.07 | 98.45 | 92.75 | 92.09 | 89.15 | 94.61 | **99.63** | **58.53** | 64.54 | 71.60 | 62.45 |
| DT | 79.76 | 85.63 | 95.56 | 52.08 | 91.06 | 76.15 | 91.41 | 79.26 | 52.61 | 58.03 | 63.59 | 59.72 |

TABLE IV
ACCURACY OF DIFFERENT METHODS ON VALIDATION NOISE DATASETS

| With Noise | Parkinson Disease | GSAD | Musk | ISOLET | Secom | Human Activity | Spambase | Protein | Kitchen | Book | Movie | Music |
|---|---|---|---|---|---|---|---|---|---|---|---|---|
| SOME-FS | **84.92** | **74.20** | **93.39** | **85.18** | 90.94 | **75.78** | **83.68** | **77.96** | **55.55** | **55.86** | **59.63** | **54.27** |
| HTSK-LN-ReLU | 78.31 | 71.29 | 90.75 | 75.18 | 89.15 | 71.39 | 80.37 | 68.61 | 54.21 | 54.82 | 59.16 | 51.62 |
| RF | 79.76 | 59.40 | 86.59 | 55.10 | **93.30** | 60.99 | 79.53 | 52.50 | 55.32 | 53.24 | 54.84 | 53.24 |
| TSK-MBGD | 79.76 | 70.74 | 91.74 | 76.52 | 86.47 | 71.74 | 81.27 | 72.41 | 53.98 | 53.37 | 57.91 | 51.77 |
| SONFIN-BN | 78.04 | 71.50 | 91.47 | 76.52 | 86.86 | 70.00 | 79.72 | 67.13 | 52.84 | 53.89 | 57.35 | 51.84 |
| RFNN | 79.63 | 68.40 | 91.59 | 79.73 | 91.89 | 70.20 | 77.29 | 73.89 | 52.15 | 53.31 | 56.32 | 50.74 |
| DT | 73.68 | 44.03 | 83.87 | 24.44 | 90.30 | 54.30 | 72.57 | 31.85 | 50.20 | 52.92 | 52.79 | 50.88 |

TABLE V
FRIEDMAN TEST ON ACCURACY ON NO NOISE DATASET

| Classifier | Mean Rank | P-values |
|---|---|---|
| SOME-FS | 1.75 | <0.01 |
| RFNN | 2.00 | |
| RF | 2.42 | |
| SONFIN | 4.75 | |
| HTSK-LN-ReLU | 5.17 | |
| TSK-MBGD | 5.25 | |
| DT | 6.67 | |

TABLE VI
HOLM'S POST HOC TEST ON NO NOISE DATASET

| Classifier | P-value | Hypothesis |
|---|---|---|
| SOME-FS | -- | -- |
| RFNN | 0.24 | Not Reject |
| RF | 0.57 | Not Reject |
| SONFIN | <0.01 | Reject |
| HTSK-LN-ReLU | <0.01 | Reject |
| TSK-MBGD | <0.01 | Reject |
| DT | <0.01 | Reject |

The above experiments confirm that SOME-FS is effective, demonstrating competitive accuracy among the state-of-the-art fuzzy classifiers and bagging methods. Furthermore, its automatic selection of rules and stable decision outputs make the proposed method applicable to various fields, producing accurate classification results. Section III.B will further test the robustness and decision stability of the proposed algorithm.

### B. Anti-Noise Performance

To evaluate the robustness of the proposed methods to uncertainty, we added standard Gaussian noise to each dataset and tested each algorithm's classification performance on the noisy data. All other experimental settings are the same as in Section III.A. The results are reported in Table IV.

The experimental results show that SOME-FS achieves the best performance on eleven datasets. RF is outperformed in only one dataset. The hypothesis test results reported in Table VII and Table VIII also confirm that SOME-FS exhibits the most robust classification performance among the seven tested classifiers. In addition, the results also show that the anti-noise ability of TSK-MGBD is significantly better than HTSK-LN-ReLU. This suggests that adding too many modules to a single TSK classifier can lead to reduced performance in noisy data.

Because SOME-FS can automatically select the rules that participate in decision-making and directly integrate the outputs of these fuzzy rules, it significantly enhances diversity. This increased diversity contributes to the method's strong robustness, which makes it highly resistant to overfitting. As a result, the model can maintain high performance across different high-dimensional datasets. It is particularly effective in handling noisy data, ensuring reliable and stable outcomes even in challenging scenarios.

TABLE VII
FRIEDMAN TEST ON ACCURACY ON NOISE DATASET

| Classifier | Mean Rank | P-Value |
|---|---|---|
| **SOME-FS** | 1.17 | <0.01 |
| TSK-MGBD | 3.33 | |
| HTSK-LN-ReLU | 3.75 | |
| SONFIN | 4.17 | |
| RFNN | 4.25 | |
| RF | 4.50 | |
| DT | 6.67 | |

TABLE VIII
HOLM'S POST HOC TEST ON NOISE DATASET

| Classifier | p-value | Hypothesis |
|---|---|---|
| **SOME-FS** | - | - |
| TSK-MGBD | <0.01 | Reject |
| HTSK-LN-ReLU | <0.01 | Reject |
| SONFIN | <0.01 | Reject |
| RFNN | <0.01 | Reject |
| RF | <0.01 | Reject |
| DT | <0.01 | Reject |

## C. Ablation Study

The key strategies of the SOME-FS proposed in this paper are divided into three basic components. These three modules interact and cooperate with each other to enhance the model's performance. We tested different combinations of these components on twelve datasets (the details of these datasets can be seen in Table II).

We evaluated four configurations of the SOME-FS:
- **SOME-FS**: Incorporates all three components.
- **Configuration Ex-RV**: Excludes the rule-based voting mechanism.
- **Configuration Ex-MSL**: Excludes the mixed structure learning module.
- **Configuration MSL-Only**: Includes only the mixed structure learning component.

The details of these configurations are reported in Table IX, and the performance metrics are shown in Figure 3 and Figure 4. The results show that SOME-FS with all components performs the best in both no noise and noisy scenarios. In no noise datasets, SOME-FS performs best on 10 datasets, while in noisy datasets, SOME-FS performs best on 9 datasets. This indicates that all three basic components in SOME-FS are essential.

TABLE IX
DIFFERENT COMBINATION OF THE THREE OPERATIONS IN SOME-FS

|  | Bagging | Mixed Structure Learning | LN-ReLU | Rule-based Voting |
|---|---|---|---|---|
| SOME-FS | √ | √ | √ | √ |
| Ex-RV | √ | √ | √ | × |
| Ex-MSL | √ | × | √ | √ |
| MSL-Only | √ | √ | × | × |

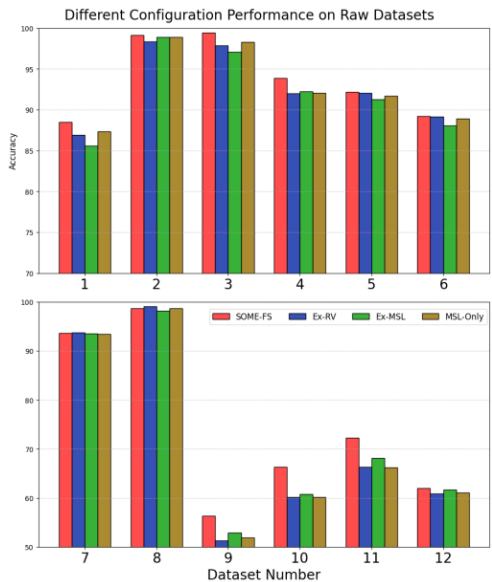

Fig. 3. Test Accuracy on the no noise dataset of the proposed SOME-FS with different operation combinations. (The red bar represents the original method combines all novel components)

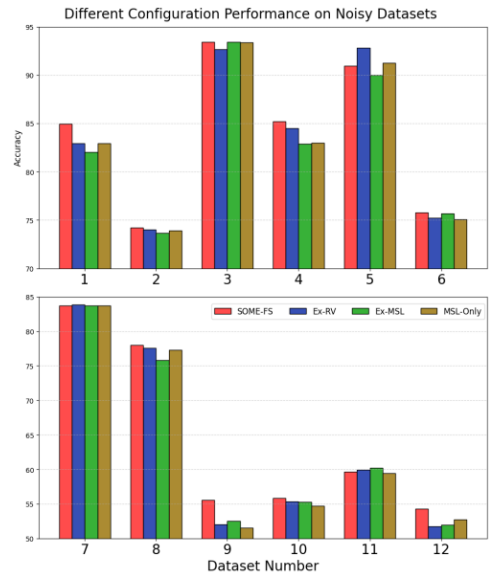

Fig. 4. Test Accuracy on the noise dataset of the proposed SOME-FS with different operation combinations. (The red bar represents the original method combines all novel components)

## D. Interpretive Analysis of Stable Rule Extraction

To better describe the unique local interpretable ability [49] of the proposed fuzzy system, we conducted a simple example using the Musk dataset (The details of the dataset can be seen in Table II). We used the dataset to train a SOME-FS model with thirty base learners and applied the mining method introduced in Section III.C to extract stable rules for a positive sample and a negative sample. This process yields some high-support rules with strong firing strength. The results are shown in Table XI and Table XII.

TABLE XI
EXTRACTED STABLE RULES ON POSITIVE SAMPLE, LINGUISTIC
FEATURE CAN BE EXPLAINED WITH MEAN AND STD

|  | Stable Rule 1 | Stable Rule 2 | Stable Rule 3 | Stable Rule 4 |
|---|---|---|---|---|
| **x17** | ex_low | ex_low | ex_low | ex_low |
| **x78** | ex_low | ex_low | ex_low |  |
| **x133** | ex_low | ex_low |  | ex_low |
| **x11** | ex_low |  |  | ex_low |
| **x18** |  | ex_low | ex_low | ex_low |
| **y** | Positive | Positive | Positive | Positive |
| **Firing Strength** | 0.138 | 0.227 | 0.189 | 0.155 |

TABLE XII

Extracted Stable Rules on Negative Sample,
Linguistic Feature can be Explained with Mean and
STD

|  | Stable Rule 1 | Stable Rule 2 | Stable Rule 3 | Stable Rule 4 |
|---|---|---|---|---|
| **x126** | high | | | |
| **x115** | high | high | high | |
| **x88** | low | low | low | low |
| **x28** | low | low | low | low |
| **x128** | | high | | |
| **x58** | | | Med | Med |
| **x5** | | | | med |
| **y** | Negative | Negative | Negative | Negative |
| **Firing Strength** | 0.465 | 0.479 | <0.001 | <0.001 |

.

Clearly, we can derive concise insights from the extracted stable rules. The linguistic rule stating that 'x17, x78, x133, x11, and x18 are extremely low' is strongly associated with a positive classification outcome. For the negative sample, the linguistic rules 'x115 is high' and 'x88 is low' are positively correlated with the classification. However, although the rule 'x58 is medium' was extracted, all associated stable rules exhibit very low firing strength, indicating minimal relevance. The interpretability is highly flexible and can be applied to medical diagnostics and fault detection scenarios to analyze the relationships between multiple core factors.

## V. Conclusion

TSK Fuzzy systems are powerful tools for handling uncertainty and imprecision in data and have been widely used in real-world classification tasks. However, their applicability to high-dimensional and noisy datasets remains a challenge. In this article, we propose the Self-Constructing Multi-Expert Fuzzy System (SOME-FS) designed to address high-dimensional data classification problems. It can effectively deal with various domain data with high dimensionality and show superior performance in handling uncertainty.

SOME-FS integrates two essential techniques, first introduced in this work:

(1) A self-constructing mechanism that automatically generates, selects and integrates relevant fuzzy rules, enhancing both robustness and adaption. Moreover, this mechanism can adapt to various data conditions and requires minimal hyperparameter tuning.

(2) A multi-expert advanced strategy combining rule selection and rule-based voting to integrate fuzzy rules within the ensemble architecture effectively. This strategy extracts the most prevalent rules to make direct decisions, greatly enhancing prediction performance and stability when dealing with high-dimensional data.

Experiments on twelve high-dimensional datasets demonstrated that SOME-FS achieves high classification accuracy and exhibits strong noise resistance. Each of the introduced techniques has its unique advantages. Integrating them can achieve superior classification performance.

However, SOME-FS has some limitations that need researching in future work. First, we shall consider extending the algorithm to other data types and exploring its scalability. Second, we aim to further enhance its performance and interpretability, potentially by incorporating additional optimization techniques or exploring alternative fuzzy rule selection methods. Additionally, we shall investigate methods to improve the computational efficiency of SOME-FS for large datasets.